# Recurrent Neural Networks for Still Images

Dmitri (Dima) Lvov, Yair Smadar, Ran Bezen
Synaptics Inc.
Herzliya, Israel
dima.lvov@synaptics.com, yair.smadar@synaptics.com, ran.bezen@synaptics.com

*Abstract*— In this paper, we explore the application of Recurrent Neural Network (RNN) for still images. Typically, Convolutional Neural Networks (CNNs) are the prevalent method applied for this type of data, and more recently, transformers have gained popularity, although they often require large models. Unlike these methods, RNNs are generally associated with processing sequences over time rather than single images. We argue that RNNs can effectively handle still images by interpreting the pixels as a sequence. This approach could be particularly advantageous for compact models designed for embedded systems, where resources are limited. Additionally, we introduce a novel RNN design tailored for two-dimensional inputs, such as images, and a custom version of Bi-Directional RNN (BiRNN) that is more memory-efficient than traditional implementations. In our research, we have tested these layers in Convolutional Recurrent Neural Networks (CRNNs), predominantly composed of Conv2D layers, with RNN layers at or close to the end. Experiments on the COCO and CIFAR100 datasets show better results, particularly for small networks.

*Keywords—RNN; recurrent; neural network; CNN; vision; still images; object deteciton; classification*

## I. INTRODUCTION

The field of computer vision has seen remarkable advancements thanks to artificial intelligence, especially using Convolutional Neural Networks (CNNs) [1]. These networks have become the go-to method for making sense of images, capturing the intricate patterns and details within them. More recently, transformers, first introduced for text inputs [2], and later also for images [3], have started to play a significant role, offering powerful capabilities but at the cost of a significantly larger memory and computational requirements. This makes them less ideal for devices with limited resources, like embedded systems.

At the same time, Recurrent Neural Networks (RNNs) have seen successful applications in the domains of speech recognition [4], natural language processing [5], video captioning [6], etc. all of which are time series. To the best of our knowledge, RNNs haven't been applied to still images. This exploration is motivated by the understanding that a sequence of pixels in an image isn't different in principle from other sequences or signals. In addition, the ability of RNN layers to have a large receptive field compared to a convolution with the same number of parameters, and hence allow replacing several convolution layers by one RNN layer, may make them more suitable for resource constrained devices, where model size and computational efficiency are crucial.

In this work we show that RNN layers can be successfully used for inference on still images, in some cases achieving better results than CNN models of a comparable size. We don't test a pure RNN network, but instead a combination of convolutional and RNN layers, in what is known as Convolutional Recurrent Neural Network (CRNN). Our main contributions include a new RNN design that's allows working with 2-dimentional inputs, called Separable RNN (SRNN), and a version of the Bi-Directional RNN (BiRNN) [7], called Weight-Shared Bidirectional RNN (WS-BiRNN), which uses less memory than the regular BiRNN. This paper also aims to open a conversation and encourage more exploration into how RNNs can be used in image analysis. Our goal is to show that there's a promising path forward in using RNNs for computer vision tasks, especially in situations were keeping things small and efficient is key.

## II. LAYER DEFINITIONS

We define the following new layer types to allow an RNN working on 2-dimensional inputs, like images.

### A. Separable RNN

*1) Regular RNN*
The regular definition of an RNN unit (layer) is,

$$h_t = \tanh(W_{ih}x_t + b_{ih} + W_{hh}h_{t-1} + b_{hh})$$

*(1)*

where $x_t \in R^{C_{in}}$ and $h_t \in R^{C_{out}}$ are the input and output features at time $t$, respectively, and $h_{t-1}$ is a hidden state of the RNN layer, stored from its previous iteration. The learnable weights are $W_{ih} \in R^{C_{out} \times C_{in}}$ and $W_{hh} \in R^{C_{out} \times C_{out}}$. The learnable biases are $b_{ih}, b_{hh} \in R^{C_{out}}$. The non-linearity chosen to be presented here is tanh, but it can as well be any other non-linear function operating elementwise. A concise notation for that is defined here as

$$H = \text{RNN}(X) \quad (2)$$

where $X = \{x_t\}_{t=1}^L$ and $H = \{h_t\}_{t=1}^L$ are 2D tensors of shape $L \times C_{in}$ and $L \times C_{out}$, respectively. The batch dimension is neglected here for clarity but is present in an actual training.

*2) Extension to 2D Inputs*

To cover the two dimensions of images, we used an approach similar to a separable convolution, in which instead of a 2D convolution two 1D convolutions are used, one on the x-axis and one on the y-axis. In a typical network designed to run on images, the input for 2-Dimentional Convolution (Conv2D) layers is a 3D tensor, with the dimensions $H \times W \times C_{in}$. This input can be treated as a 2-dimensional sequence of feature vectors,

$$X = \{x_{jk}\}_{j=1..H, k=1..W} \quad x_{jk} \in R^{C_{in}} \quad (3)$$

and the regular RNN layer is well-defined when operating independently on every row,

$$h_{jk} = \{\text{RNN}(X)\}_{jk} = \tanh(W_{ih} x_{jk} + b_{ih} + W_{hh} h_{j,k-1} + b_{hh}). \quad (4)$$

Note that the same weights and biases are used for every row, as would be the case in a separable convolution. This produces an intermediate result

$$H_1 = \text{RNN}_1(X) \quad (5)$$

The second RNN layer operates on the columns of $H_1$, or the rows of $H_1^T$

$$H_2 = \text{RNN}_2(H_1^T)^T \quad (6)$$

where the transpose operation operates on the height and width dimensions,

$$\{X^T\}_{jk} = \{X\}_{kj} = x_{kj}. \quad (7)$$

Finally, a Separable RNN (SRNN), is defined as

$$\text{SRNN}(X) = \text{RNN}_2(\text{RNN}_1(X)^T)^T \quad (8)$$

Note that although it is possible to use the same weights for the rows and the columns, we use $\text{RNN}_1$ and $\text{RNN}_2$ which have different weights. Fig. 1(a-c) present a visual example of RNN and SRNN operating on an image. The input image contains two impulses at different locations, and the RNN and SRNN outputs show the impulse responses of those.

*B. Weight-Shared Bidirectional RNN*

Convolution layers have a symmetric receptive field around the center of their kernel. In contrast, a basic RNN layer has a one-sided receptive field, meaning that the calculation of $h_t$, through the recursive usage of $h_{t-1}$, includes only the previous inputs, $\{x_{t-k}: k \in N\}$. A common solution for that is using a Bidirectional RNN (BiRNN), which contains two RNN layers, one operating on the input sequence as in (1, and the other operates on the same sequence in a reverse direction. To reduce the number of parameters, we propose a Weight-Shared Bidirectional RNN (WS-BiRNN) layer, which is similar to the BiRNN except that the same weights are used for both directions,

$$\text{WS-BiRNN}(X) = \text{RNN}_1(X) + \text{Flip}\left(\text{RNN}_1(\text{Flip}(X))\right) \quad (9)$$

where the Flip operation reverses the order of the sequence $X$, or in case of 2 dimensions, the order of the rows of $X$,

$$\{\text{Flip}(X)\}_{jk} = \{X\}_{j, W+1-k} \quad (10)$$

This formulation forces the effective weights multiplying $\{x_t\}$ in the calculation of $h_t$ to be symmetric around $t$. The extension to 2 dimensions similar to SRNN is denoted as Separable Weight-Shared Bidirectional RNN (SWS-BiRNN), and defined by,

$$\text{SWS-BiRNN}(X) = \text{WS-BiRNN}_2(\text{WS-BiRNN}_1(X)^T)^T \quad (11)$$

Fig. 1 (d, e) show the responses of the WS-BiRNN and SWS-BiRNN to the two impulses at the input. The responses show a wide field of view of these layers. Notice that they are most strong at the same row and column of the input impulse.

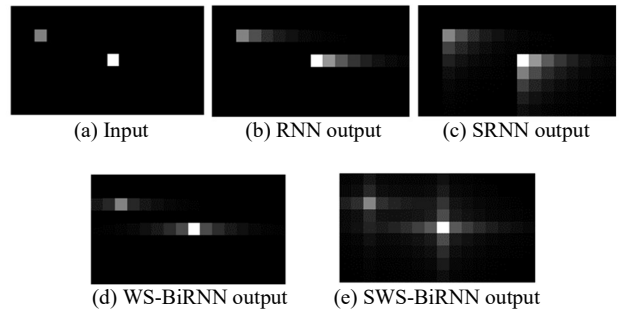

(a) Input    (b) RNN output    (c) SRNN output

(d) WS-BiRNN output    (e) SWS-BiRNN output

Figure 1: Operation illustration of the different defined RNN variants on a 16×9 picture with 1 channel.

## C. Resources Usage of RNN Compared to CNN

### 1) Number of Parameters

Denoting the number of output channels of the first RNN in a SRNN, which is also the number of input channels for the second RNN, by $C_{mid}$, the number of parameters in a SRNN or a SWS-BiRNN layer is,

$$\text{SWS-BiRNN}_{\text{parameters}} = C_{in} \cdot C_{mid} + C_{mid}^2 + 2C_{mid} \\ + C_{mid} \cdot C_{out} + C_{out}^2 + 2C_{out}. \quad (12)$$

The two bias vectors, $b_{ih}$ and $b_{hh}$, could be combined into a single vector of the same size, but here we follow the implementation of PyTorch, which holds them separately. The number of parameters in a Conv2D layer is,

$$\text{Conv2D}_{\text{parameters}} = C_{in} \cdot C_{out} \cdot k^2 + C_{out}, \quad (13)$$

were $k$ is the kernel size. Since $C_{mid}$ and $C_{out}$ have square terms in the SWS-BiRNN case, in cases when $C_{mid}$ or $C_{out}$ are considerably larger than $C_{in}$, the number of parameters in a SWS-BiRNN may be larger than in a Conv2D layer with a small kernel. But when the number of output and input channels is similar, $C_{in} = C_{mid} = C_{out} = C \gg 1$,

$$\text{SRNN}_{\text{parameters}} = 4C^2 + 4C \quad (14)$$

$$\text{Conv2D}_{\text{parameters}} = k^2 C^2 + C \quad (15)$$

$$\frac{\text{SWS-BiRNN}_{\text{parameters}}}{\text{Conv2D}_{\text{parameters}}} \approx \frac{4}{k^2} \quad (16)$$

which is usually smaller than 1, except for a pointwise convolution.

It may be noted that using a separable convolution layer with a small kernel may be similar to SWS-BiRNN in the number of parameters.

### 2) Number of Multiplications

In embedded systems, it is common to use the number of Multiply and Accumulate (MAC) operations in an algorithm as a good approximation of its computational complexity. Here we'll also count the number of additions as MACs, although it's usually much smaller. The number of MACs of a Conv2D for an input of size $H \times W \times C_{in}$ is,

$$\text{Conv2D}_{\text{MACs}} = (k^2 \cdot C_{in} \cdot C_{out} + C_{out}) \cdot H \cdot W, \quad (17)$$

while for a SRNN it is,

$$\text{SRNN}_{\text{MACs}} = (C_{in} C_{mid} + C_{mid}^2 + 2C_{mid} + C_{mid} C_{out} + C_{out}^2 \\ + 2C_{out}) \cdot H \cdot W. \quad (18)$$

As with the number of parameters, $C_{mid}$ and $C_{out}$ have square terms in the SRNN case. Thus, in cases when $C_{mid}$ or $C_{out}$ are considerably larger than $C_{in}$, the number of MACs will be higher in a SRNN layer than in a Conv2D layer with a small kernel. But when the number of output and input channels is similar, $C_{in} = C_{mid} = C_{out} = C$, the ratio of MACs is,

$$\frac{\text{SRNN}_{\text{MACs}}}{\text{Conv2D}_{\text{MACs}}} = \frac{2(C^2 + C^2 + 2C)}{k^2 C^2 + C} = \frac{4C + 4}{k^2 C + 1}. \quad (19)$$

assuming that $C$ is much larger than 1, the ratio of MACs is $4/k^2$, which is usually smaller than 1. As for the SWS-BiRNN layer, it has double the MACs compared to SRNN, so its ratio of MACs compared to Conv2D is $8/k^2$, which is usually close to 1.

## D. Replacing Several Conv2D Layers with one SRNN

The bigger advantage of SRNN or SWS-BiRNN over convolution layers may lie in their ability of getting a large receptive field with no need in increasing the number of parameters, or the kernel size. In small networks, in order to capture big or close objects in an image, either a Conv2D layer with a very big kernel, or multiple such layers with small kernels must be used. One SRNN or SWS-BiRNN layer can potentially replace several convolution layers, thus considerably reducing the number or parameters and multiplications.

In addition, when using parallel execution on a Neural Processing Unit (NPU), except for the MACs count there is an overhead of starting the processing of every layer, which in some cases is related to rearrangement and duplication of the input values, allowing the parallel operation. For small networks / layers, this overhead is not negligible, thus replacing several Conv2D layers by one SRNN or SWS-BiRNN layer can considerably reduce the inference time.

## III. EXPERIMENTS AND RESULTS

We conducted experiments focusing on two tasks: object detection and classification. For object detection, we utilized a CNN network trained on the COCO dataset [8], developed in-house. This network outperforms YOLOv5 [9] in object detection efficiency on embedded devices with very limited resources, attributed to its optimal balance between the number of parameters and Tensor arena size. The architecture of the CNN model features a Backbone comprising a convolution layer followed by 15 modified Inverted Residual blocks [10], in which the first bottleneck and the depth-wise separable convolutions were replaced by a regular 2D convolution (See Fig. 2). In addition, it includes a Neck with 9 convolution layers and 5 heads, emanating from different points within the Neck, each consisting of 1-4 convolution layers predicting for 3 anchor boxes each. This model specifically detects objects of the 'person' class. For classification, we trained CIFAR100 on Resnet18 [11] with a width-multiplier of 0.2, incorporating 4 Residual blocks (See Table 6), a convolution layer, and one fully connected layer for the final classification. The basic residual blocks are presented in Fig. 3.

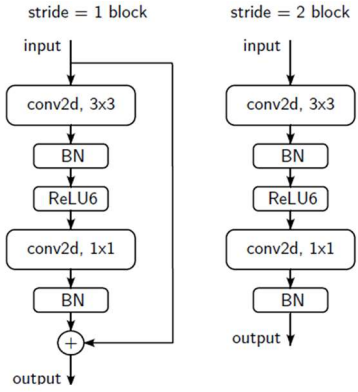

Figure 2: A modified version of the Inverse residual block from MobileNetV2, producing a smaller tensor arena and better computational efficiency on a Neural Processing Unit (NPU).

TABLE 1: OBJECT DETECTION NETWORK DETAILS FOR 256 X 256 INPUTS

| Index | From | n | Input | Layer | Params | Exp | s | CNN | CRNN |
|---|---|---|---|---|---|---|---|---|---|
| 0 | -1 | 1 | 256×256×3 | Conv2D | 135 | - | 2 | 3×3 | |
| 1 | 0 | 1 | 128×128×5 | InvRes | 1160 | 20 | 2 | ✓ | |
| 2 | 1 | 1 | 64×64×10 | InvRes | 2060 | 20 | 1 | ✓ | |
| 3 | 2 | 1 | 64×64×10 | InvRes | 13358 | 120 | 2 | ✓ | |
| 4 | 3 | 2 | 32×32×19 | InvRes | 43852 | 114 | 1 | ✓ | |
| 5 | 4 | 1 | 32×32×19 | InvRes | 22854 | 114 | 2 | ✓ | |
| 6 | 5 | 6 | 16×16×27 | InvRes | 176580 | 108 | 1 | ✓ | |
| 7 | 6 | 1 | 16×16×27 | InvRes | 30090 | 108 | 2 | ✓ | |
| 8 | 7 | 2 | 8×8×33 | InvRes | 87780 | 132 | 1 | ✓ | |
| 9 | 8 | 1 | 8×8×33 | Conv2D | 16632 | - | 1 | 3×3 | |
| 10 | 4 | 1 | 32×32×19 | Conv2D | 703 | - | 1 | 1×1 | |
| 11 | 6 | 1 | 16×16×27 | Conv2D | 513 | - | 1 | 1×1 | |
| 12 | 9 | 1 | 8×8×56 | Conv2D | 1064 | - | 1 | 1×1 | |
| 13 | 12 | 1 | 8×8×19 | Conv2D | 361 | - | 1 | 1×1 | |
| 14 | 12 | 1 | 8×8×19 | Conv2D | 380 | - | 1 | 1×1 | |
| 15 | 14 | 1 | 8×8×19 | Conv2D | 3268 | - | 2 | 3×3 | |
| 16 | 15 | 1 | 4×4×19 | Conv2D | 380 | - | 1 | 1×1 | |
| 17 | 16 | 1 | 4×4×19 | Conv2D | 3268 | - | 1 | 1×1 | |
| 18 | 10 | 4/1 | 32×32×37 | Conv2D / SWS-BiRNN | 49432 / 27990 | - | 1 | 3×3 | 1×1 |
| 19 | 11 | 4/1 | 16×16×19 | Conv2D / SWS-BiRNN | 13072 / 14170 | - | 1 | 3×3 | 1×1 |
| 20 | 14 | 4/1 | 8×8×19 | Conv2D / SWS-BiRNN | 13072 / 14170 | - | 1 | 3×3 | 1×1 |
| 21 | 15 | 2/1 | 4×4×19 | Conv2D / SWS-BiRNN | 6536 / 14170 | - | 1 | 3×3 | 1×1 |
| 22 | 17 | 1 | 2×2×19 | Conv2D / SWS-BiRNN | 3268 / 14170 | - | 1 | 3×3 | 1×1 |

TABLE 2: OBJECT DETECTION NETWORK DETAILS FOR 480 X 640 INPUTS.

| Index | From | n | Input | Layer | Params | Exp | s | CNN | CRNN |
|---|---|---|---|---|---|---|---|---|---|
| 0 | -1 | 1 | 480×640×3 | Conv2D | 216 | - | 2 | 3×3 | |
| 1 | 0 | 1 | 240×320×8 | InvRes | 2878 | 32 | 1 | ✓ | |
| 2 | 1 | 1 | 120×160×15 | InvRes | 4590 | 30 | 1 | ✓ | |
| 3 | 2 | 1 | 120×160×15 | InvRes | 29938 | 180 | 2 | ✓ | |
| 4 | 3 | 2 | 60×80×29 | InvRes | 101732 | 174 | 1 | ✓ | |
| 5 | 4 | 1 | 60×80×29 | InvRes | 49634 | 174 | 2 | ✓ | |
| 6 | 5 | 6 | 30×40×22 | InvRes | 117650 | 88 | 1 | ✓ | |
| 7 | 6 | 1 | 30×40×22 | InvRes | 19580 | 88 | 2 | ✓ | |
| 8 | 7 | 2 | 15×20×22 | InvRes | 39160 | 88 | 1 | ✓ | |
| 9 | 8 | 1 | 15×20×22 | Conv2D | 16830 | - | 1 | 3×3 | |
| 10 | 4 | 1 | 60×80×29 | Conv2D | 1653 | - | 1 | 1×1 | |
| 11 | 6 | 1 | 30×40×22 | Conv2D | 638 | - | 1 | 1×1 | |
| 12 | 9 | 1 | 15×30×85 | Conv2D | 2465 | - | 1 | 1×1 | |
| 13 | 12 | 1 | 15×30×29 | Conv2D | 841 | - | 1 | 1×1 | |
| 14 | 12 | 1 | 15×30×29 | Conv2D | 870 | - | 1 | 1×1 | |
| 15 | 14 | 1 | 15×30×29 | Conv2D | 7598 | - | 2 | 3×3 | |
| 16 | 15 | 1 | 8×15×29 | Conv2D | 870 | - | 1 | 1×1 | |
| 17 | 16 | 1 | 8×15×29 | Conv2D | 7598 | - | 2 | 3×3 | |
| 18 | 10 | 4/1 | 60×80×57 | Conv2D / SWS-BiRNN | 117192 / 16640 | - | 1 | 3×3 | 1×1 |
| 19 | 11 | 4/1 | 30×40×29 | Conv2D / SWS-BiRNN | 30392 / 4224 | - | 1 | 3×3 | 1×1 |
| 20 | 13 | 4/1 | 15×30×29 | Conv2D / SWS-BiRNN | 30392 / 4224 | - | 1 | 3×3 | 1×1 |
| 21 | 15 | 2/1 | 8×15×29 | Conv2D / SWS-BiRNN | 15196 / 4224 | - | 1 | 3×3 | 1×1 |
| 22 | 17 | 1 | 4×8×29 | Conv2D / SWS-BiRNN | 7598 / 4224 | - | 1 | 3×3 | 1×1 |

TABLE 3: OBJECT DETECTION RESULTS WITH 480X640 INPUTS

| Model | Parameters | Tensor arena | Total Memory | F1 Score | AP50 |
|---|---|---|---|---|---|
| CNN | 605K | 1.63M | 2.23M | **68.0** | **70.1** |
| CRNN | 522K | 1.63M | 2.15M | 67.4 | 68.5 |

TABLE 4: OBJECT DETECTION RESULTS WITH 480X640 INPUTS

| Model | Parameters | Tensor arena | Total Memory | F1 Score | AP50 |
|---|---|---|---|---|---|
| CNN | 493K | 819K | 1.31M | 55.5 | 53.1 |
| CRNN | 493K | 819K | 1.31M | **56.3** | **53.8** |
| YOLOv5 | 101K | 1.24M | 1.34M | 44.4 | 39.7 |

To assess the SWS-BiRNN's effectiveness on COCO, we replaced each CNN model head with an SWS-BiRNN layer, adjusting the channel count to align the parameter totals of both networks. Details can be found in Tables 1 and 2. This modified network, a CRNN, was evaluated with two different input dimensions, 256x256 and 480x640, which is a significant factor in determining Tensor arena (work memory) size. For comparative analysis, we also trained YOLOv5 with a similar total memory (parameters + tensor arena) as the compact CRNN. The results, presented in Table 3, indicate that the larger CNN model, with a memory footprint of 2.15M-2.23M, marginally outperforms the CRNN in F1 Score and AP50 metrics. Conversely, as shown in Table 4, the CRNN excels over both YOLOv5 and the CNN model in a smaller configuration, with memory allocations between 1.31M-1.34M. These results suggest that the SWS-BiRNN's may be beneficial particularly in networks with fewer parameters. In our experiments we also tried variants of the SWS-BiRNN based not on the basic RNN but instead on LSTM [12] or GRU [13]. However, those led to inferior results.

For CIFAR100, we evaluated the SWS-BiRNN by replacing the last Residual block of ResNet18 with an SWS-BiRNN layer, which contains the most parameters in the model, creating the ResNet18-RNN. Additionally, to compare with a similarly sized model, we adjusted another two models, ResNet18-thin model to reduce the parameter count in its last Residual block and ResNet18-short, with single Conv2D layer instead of the last Residual block to compare the performance of SWS-BiRNN and Conv2D with the same number of channels. For comprehensive information on the model architectures, refer to Table 5. According to Table 6, the ResNet18-RNN exhibits lower accuracy relative to the larger ResNet18. Yet, against the smaller variants of ResNet18-thin and ResNet18-short, with an equal or smaller parameter count, it demonstrates superior accuracy.

## IV. CONCLUSIONS

The exploration into the adaptability of RNN layers, traditionally favored for time series, has demonstrated their efficacy in processing still images as well. The strategic replacement of multiple CNN layers with a solitary SWS-

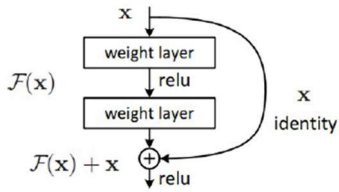

Figure 3: The original residual block, used here for the CIFAR100 experiments.

TABLE 5: CLASSIFICATION NETWORKS STRUCTURE

| Layer name | Output | #Parameters | ResNet18 | ResNet18-RNN | ResNet18-thin | ResNet18-short |
|---|---|---|---|---|---|---|
| conv1 | 32×32 | 324 | | 3 × 3, 12 | | |
| conv2_x | 32×32 | 5,280 | | 3 × 3, 12 ×2 <br> 3 × 3, 12 | | |
| conv3_x | 16×16 | 20,125 | | 3 × 3, 25 ×2 <br> 3 × 3, 25 | | |
| conv4_x | 8×8 | 83,487 | | 3 × 3, 51 ×2 <br> 3 × 3, 51 | | |
| conv5_x | 4×4 | 333,948 / 44,288 | 3 × 3, 102 ×2 <br> 3 × 3, 102 | - | 3 × 3, 32 ×2 <br> 3 × 3, 32 | - |
| conv6 | 4×4 | 46,920 | - | - | - | 3 × 3, 102 |
| SWS-BiRNN | 8×8 | 36,822 | - | 1 × 1, 102 <br> 1 × 1, 102 | - | - |
| AvgPool | 1×1 | - | 4 × 4, 102 | 8 × 8, 102 | 4 × 4, 32 | 4 × 4, 102 |
| FC | 1×1 | 10,300 / 3,300 | 102 × 100 | 102 × 100 | 32 × 100 | 102 × 100 |

TABLE 6: CLASSIFICATION TASK RESULTS

| Model | Parameters | Tensor arena | Total Memory | Accuracy |
|---|---|---|---|---|
| ResNet18 | 453K | 24K | 477K | **61.4** |
| ResNet18-RNN | 156K | 24K | 180K | 59.0 |
| ResNet18-thin | 156K | 24K | 180K | 57.1 |
| ResNet18-short | 166K | 24K | 190K | 58.1 |

BiRNN layer emerges as particularly beneficial within compact network structures. This approach renders it a highly suitable option for deployment on edge devices, where resource constraints are prevalent. Conversely, in scenarios involving larger models, the utilization of CNNs continues to present a more advantageous route. This distinction underscores the importance of tailoring the network architecture to the specific requirements of the deployment, balancing computational efficiency and model performance.

## V. FUTURE WORK: 2-DIMENSIONAL RNN

*1) 2-Deminisional RNN (RNN2D)*

A separable convolution is a private case of a Conv2D, able to represent only 2D kernels with linearly dependent rows, and columns. Similar to that, the SRNN is only able to represent exponentially decaying weighting of the input which has its principal directions on the x and y axes. A more general, 2-Deminsional RNN operation may be defined as,

$$h_{jk} = \tanh(W_{ih}x_{jk} + W_{hk}h_{j,k-1} + W_{hj}h_{j-1,k} + W_{jk}h_{j-1,k-1} + b)$$
(20)

allowing the principal direction of the input weighting to be diagonal, or at any angle in the $xy$ plane, when combined with a bidirectional operation.

*2) Depthwise Separable (DS) RNN*

In addition, the principle of Depthwise Separable (DS) convolution can be applied to RNN, reducing the computational and memory requirements. One of the possible ways to define such a thing can be

$$h_{jk} = \tanh(W_{ih} \odot x_{jk} + W_a \odot h_{j,k-1} + W_b \odot h_{j-1,k} + \cdots + b)$$
(21)

$$y_{jk} = C \cdot h_{jk} : C \in R^{C_{out} \times C_{in}}$$
(22)

Where $h, w, x \in R^{C_{in}}$, $\odot$ denotes an element-wise multiplication, and $C \in R^{C_{out} \times C_{in}}$ is a $1 \times 1$ convolution kernel. The recurrent operation is made independently on every channel and followed by a $1 \times 1$ convolution.

However, these directions require further development of the mathematical formulation, and experimentation, which we haven't conducted yet. We leave that for possible future work.

REFERENCES

[1] Le Cun Y. *et al.*, "Handwritten Digit Recognition: Applications of Neural Net Chips and Automatic Learning," in *Neurocomputing*, J. Soulié Françoise Fogelman and Hérault, Ed., Berlin, Heidelberg: Springer Berlin Heidelberg, 1990, pp. 303–318.

[2] V. Ashish *et al.*, "Attention is All you Need," *Adv Neural Inf Process Syst*, pp. 5998–6008, 2017.

[3] A. Dosovitskiy *et al.*, "An Image is Worth 16x16 Words: Transformers for Image Recognition at Scale," Oct. 2020, Accessed: Mar. 14, 2024. [Online]. Available: http://arxiv.org/abs/2010.11929

[4] Y. Bengio and Y. Bengio, "Artificial neural networks and their application to sequence recognition." McGill University, 1991. Accessed: Mar. 14, 2024. [Online]. Available: https://escholarship.mcgill.ca/concern/theses/qv33rx48q

[5] W. Yin, K. Kann, M. Yu, and H. Schütze, "Comparative Study of CNN and RNN for Natural Language Processing," Feb. 2017, Accessed: Mar. 14, 2024. [Online]. Available: http://arxiv.org/abs/1702.01923

[6] L. Gao, Z. Guo, H. Zhang, X. Xu, and H. T. Shen, "Video Captioning With Attention-Based LSTM and Semantic Consistency," *IEEE Trans Multimedia*, vol. 19, no. 9, pp. 2045–2055, Sep. 2017, doi: 10.1109/TMM.2017.2729019.

[7] M. Schuster and K. K. Paliwal, "Bidirectional recurrent neural networks," *IEEE Transactions on Signal Processing*, vol. 45, no. 11, pp. 2673–2681, 1997, doi: 10.1109/78.650093.

[8] T.-Y. Lin *et al.*, "Microsoft COCO: Common Objects in Context," Springer, Cham, 2014, pp. 740–755. doi: 10.1007/978-3-319-10602-1_48.

[9] "GitHub - ultralytics/yolov5: YOLOv5 🚀 in PyTorch > ONNX > CoreML > TFLite." Accessed: Dec. 26, 2021. [Online]. Available: https://github.com/ultralytics/yolov5

[10] M. Sandler, A. Howard, M. Zhu, A. Zhmoginov, and L.-C. Chen, "MobileNetV2: Inverted Residuals and Linear Bottlenecks," Jan. 2018, Accessed: Nov. 14, 2019. [Online]. Available: http://arxiv.org/abs/1801.04381

[11] Kaiming He, Xiangyu Zhang, Shaoqing Ren, Jian Sun. "Deep Residual Learning for Image Recognition" Dec 2015, Accessed: Dec. 10, 2015. [Online]. Available: https://arxiv.org/abs/1512.03385

[12] A. Graves, "Long Short-Term Memory," Springer, Berlin, Heidelberg, 2012, pp. 37–45. doi: 10.1007/978-3-642-24797-2_4.

[13] R. Dey and F. M. Salem, "Gate-variants of Gated Recurrent Unit (GRU) neural networks," in *2017 IEEE 60th International Midwest Symposium on Circuits and Systems (MWSCAS)*, IEEE, Aug. 2017, pp. 1597–1600. doi: 10.1109/MWSCAS.2017.8053243.